\documentclass[letterpaper, 10 pt, conference]{ieeeconf}  

\IEEEoverridecommandlockouts                              

\usepackage{booktabs}
\usepackage{hyperref}
\usepackage[pdftex]{graphics} 
\usepackage{epsfig} 

\usepackage{makecell}
\usepackage{mathtools}
\usepackage{mathrsfs} 
\usepackage{times} 
\usepackage[maxfloats=256]{morefloats}
\usepackage{amsmath,amssymb,amsfonts,bm,amscd}

\usepackage{amsthm} 
\usepackage[nameinlink]{cleveref}
\usepackage{algpseudocode}
\usepackage{setspace}
\usepackage{fancyhdr}
\usepackage{algorithm} 
\usepackage{psfrag}
\usepackage{cite}
\usepackage{makecell}
\usepackage{url}
\usepackage{float}
\usepackage{soul}
\usepackage{color}
\usepackage{multirow}
\usepackage{color, xcolor}
\usepackage{etoolbox}    
\usepackage[T1]{fontenc}
\usepackage{epsfig,subfigure}
\allowdisplaybreaks[4]
\title{\LARGE \bf
Occlusion-Aware Consistent Model Predictive Control for 
Robot Navigation in Occluded Obstacle-Dense Environments
}
\newcommand{\argmin}[1]{\underset{#1}{\displaystyle\operatorname*{ \text{argmin}}}\;}
\newtheorem{assumption}{\textbf{Assumption}}
\newtheorem{remark}{Remark}

\makeatletter
\newcommand\fs@spaceruled{\def\@fs@cfont{\bfseries}\let\@fs@capt\floatc@ruled
  \def\@fs@pre{\vspace{3mm}\hrule height.1pt depth0pt \kern2pt}%
  \def\@fs@post{ \kern2pt\hrule\relax}%
  \def\@fs@mid{\kern2pt\hrule\kern2pt}%
  \let\@fs@iftopcapt\iftrue}
\makeatother

\floatstyle{spaceruled}
\restylefloat{algorithm}

\author{
Minzhe Zheng, Lei Zheng, Lei Zhu, and Jun Ma, \textit{Senior Member, IEEE}
\thanks{This work was supported in part by the Guangdong Basic and Applied Basic Research Foundation under Grant 2025A1515011812; and in part by the Guangdong provincial project under Grant 2023QN10Z006. Minzhe Zheng and Lei Zheng contributed equally to this work. \textit{(Corresponding Author: Jun Ma.)}}
\thanks{Minzhe Zheng, Lei Zheng, Lei Zhu, and Jun Ma are with the Robotics and Autonomous Systems Thrust, The Hong Kong University of Science and Technology (Guangzhou), Guangzhou 511453, China (e-mail: mzheng615@connect.hkust-gz.edu.cn; lzheng135@connect.hkust-gz.edu.cn; leizhu@hkust-gz.edu.cn; jun.ma@ust.hk).}
\thanks{This work has been submitted to the IEEE for possible publication.
Copyright may be transferred without notice, after which this version may
no longer be accessible.}
}

\begin{document}

\maketitle
\pagestyle{empty}

\begin{abstract}
Ensuring safety and motion consistency for robot navigation in occluded, obstacle-dense environments is a critical challenge. In this context, this study presents an occlusion-aware Consistent Model Predictive Control (CMPC) strategy. 
To account for the occluded obstacles, it incorporates adjustable risk regions that represent their potential future locations. Subsequently, dynamic risk boundary constraints are developed online to enhance safety.
Based on these constraints, the CMPC constructs multiple locally optimal trajectory branches (each tailored to different risk regions) to strike a balance between safety and performance. A shared consensus segment is generated to ensure smooth transitions between branches without significant velocity fluctuations, preserving motion consistency. To facilitate high computational efficiency and ensure coordination across local trajectories, we use the alternating direction method of multipliers (ADMM) to decompose the CMPC into manageable sub-problems for parallel solving. The proposed strategy is validated through simulations and real-world experiments on an Ackermann-steering robot platform. 
The results demonstrate the effectiveness of the proposed CMPC strategy through comparisons with baseline approaches in occluded, obstacle-dense environments.
\end{abstract}

\section{Introduction}
Ensuring safe navigation and consistent motion for mobile robots in occluded, obstacle-dense environments is a critical challenge~\cite{10819000}. One of the key underlying factors to this concern is the partial observability of such environments due to occlusions~\cite{10342223, 9982059}. In this context, robots relying on onboard perception, which typically operate on line-of-sight principles, are unable to detect obstacles hidden by occlusions~\cite{10305287,chen2025robotnavigationunknowncluttered}. This limitation increases the risk of collisions, particularly in obstacle-dense environments. Additionally, the sudden emergence of occluded obstacles can lead to abrupt velocity changes for the robot, compromising motion consistency and stability.

\begin{figure}[tp]
    \centering
\includegraphics[width=0.95\columnwidth]{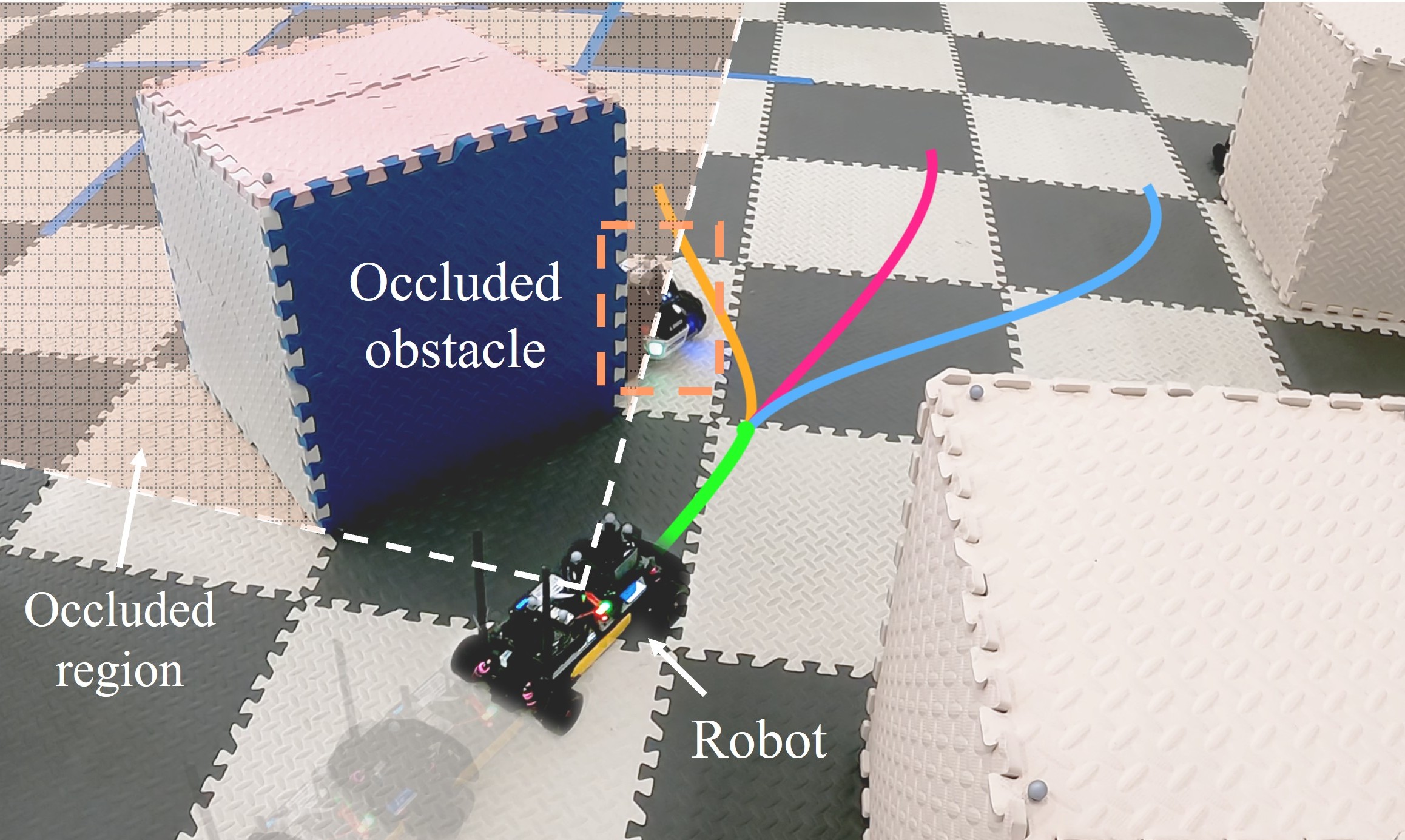}
    \caption{Robot navigation in an occluded, obstacle-dense environment using the proposed occlusion-aware CMPC. The occlusion-aware CMPC generates three trajectories (orange, red, and blue) with different considerations of risk regions. All trajectories share an initial consensus segment (green) to enable smooth transitions between trajectories and ensure motion consistency. } 
    \label{fig:descript}
    \vspace{-1em}
\end{figure} 

To ensure safe navigation in obstacle-dense environments, model predictive control (MPC) facilitates obstacle avoidance through receding-horizon optimization, which systematically accounts for system dynamics and environmental constraints to achieve proactive collision avoidance~\cite{9612725,10160857}. To construct obstacle avoidance constraints, accurate prediction of obstacle states is essential for this strategy. Hence, robots must account for potential states of occluded obstacles to guarantee safety in occluded scenarios~\cite{10819000}. However, precise state estimation of fully occluded obstacles remains infeasible~\cite{ranaraja2024occlusion}. An alternative way involves predicting the risk regions of these obstacles, which reduces planning complexity and enhances safety margins~\cite{10305287}. For instance, OA-MPC~\cite{10819000} adopts worst-case assumptions to ensure safety. This approach theoretically covers potential states of occluded obstacles. However, the generated trajectory tends to be overly conservative~\cite{9561664}. Moreover, a large deviation from the reference path compromises motion consistency~\cite{zheng2025occlusion}. 

To enhance motion consistency, recent studies have proposed scenario-adaptive planning strategies~\cite{10706022,zheng2025barrier}. In these approaches, multiple homotopic locally optimal trajectories are generated for autonomous vehicles. Although consistent parameters are used within the decision-making module to select the optimal trajectory and maintain motion consistency, each trajectory is optimized separately. Essentially, the lack of coordination can lead to velocity fluctuations, particularly in occluded, obstacle-dense settings.
Alternatively, branch MPC (BMPC)~\cite{10234564,10737675} introduces a shared common consensus segment across branches, ensuring smooth transitions between trajectory branches and maintaining motion consistency. Similarly, partially observable Markov decision processes (POMDP)~\cite{9899480} ensures environmental consistency through belief state updates~\cite{10563192}. However, these approaches encounter scalability challenges, as branch proliferation significantly increases the number of optimization variables, leading to intractable computation. To address this challenge, recent works have employed the alternating direction method of multipliers (ADMM), which decomposes the original problem into low-dimensional sub-problems to enable real-time performance without compromising safety~\cite{zheng2024saferealtimeconsistentplanning,10610987,10036019}. 
Nonetheless, despite the strong computational performance of ADMM,
a more definite strategy is
still lacking on how to extend its usage to address the challenges posed by occluded, obstacle-dense environments.

In this study, we introduce a novel occlusion-aware Consistent Model Predictive Control (CMPC) strategy for mobile robots navigating in occluded, obstacle-dense environments. The CMPC optimizes several locally optimal trajectories in parallel to account for potential occlusion risks, as shown in Fig.~\ref{fig:descript}. Subsequently, an ADMM-based optimization strategy is used to decompose the optimization problem into parallel sub-problems to facilitate computational efficiency. 
The main contributions of this paper are summarized as follows:
\begin{itemize} 
    \item
    We develop a computationally efficient CMPC optimization approach for safe navigation of mobile robots in occluded, obstacle-dense environments. It leverages the ADMM to decompose the optimization problem into sub-problems and solve them in parallel. This strategy enables the robot to achieve real-time planning while ensuring safety.
    
    \item We introduce a consistent motion planning strategy under occlusion, where multiple trajectory branches share a consensus segment to ensure motion consistency. This strategy facilitates adjustable risk region configurations as dynamic risk boundary constraints in trajectory generation to strike a balance between safety and performance. With this strategy, the robot exhibits less conservative behavior in occluded, obstacle-dense environments with enhanced motion consistency.

    \item We validate our strategy through simulations and real-world experiments conducted on an Ackermann-steering mobile robot platform in occluded, obstacle-dense environments. The results and comparison with baselines demonstrate the effectiveness of the proposed approach in ensuring safe and consistent navigation in occluded, obstacle-dense environments. 
\end{itemize}

\section{Related Work}\label{Relate work}

\subsection{Reachability Analysis}
Reachability analysis is an efficient approach in occlusion-aware planning, as demonstrated in extensive studies \cite{9827171, 10305287,8793557, 8569332}. It creates a reachable set of agents in the occluded regions to construct constraints for planning. In \cite{5109666}, the impact of environmental visibility on mobile robot navigation is considered to ensure safety. An MPC strategy is proposed to apply risk fields from reachable sets for automated vehicles~\cite{10495173}. However, reachability analysis typically focuses on worst-case scenarios, leading to overly conservative behaviors. Additionally, predicting obstacle states based on rules of on-road scenarios~\cite{9827171,10305287} is not applicable to occluded, obstacle-dense environments encountered by mobile robots.

\subsection{POMDP}
POMDP has been widely adopted in occlusion-aware planning for its ability to handle partially observable scenarios~\cite{9899480,10610876}. It leverages a probability distribution in the prediction of the agents' future states for effective planning. For instance, works \cite{9197064, 9564424} leverage the POMDP framework to address navigation and decision-making problems in dynamic environments with partial occlusions. \cite{10563192} further incorporates a belief state updating module to predict the scenarios more precisely for effective planning. While these approaches can tackle partially observed obstacles in structured road environments, their high computational complexity poses challenges for real-time planning. Additionally, obtaining reliable probability distributions is difficult in unstructured environments with dense occlusion, potentially leading to planning failures. 

\subsection{Contingency Planning}
Contingency planning enhances safety and motion consistency by providing multiple contingency paths to address uncertainties in the future states of dynamic obstacles~\cite{6497657,9729171, 10234564, 10178332}. \cite{packer2022is} presents a learning-based method for predicting the states of partially observed agents and motion planning on road. However, existing approaches typically require substantial computational resources to determine policies based on assumptions about future states. Additionally, it typically makes rule-based assumptions about the future states in on-road scenarios, or one-to-one interaction between agents~\cite{zheng2025occlusion}. These future state assumptions may not be reliable in general scenarios. 

In this study, we follow the concept of reachability analysis to model the risk regions where occluded obstacles may appear in the future. We predict multiple configurations of risk regions while avoiding explicit hypotheses about obstacle states. To address computational complexity, we propose an occlusion-aware CMPC strategy combined with an ADMM-based optimization process. This strategy ensures safety, efficiency, and adaptability in occluded, obstacle-dense environments.
\section{Problem Statement}\label{problems and pre} \label{problem statement}
In this study, we consider a robot modeled by the simplified unicycle model~\cite{Lynch_Park_2017}, with its state $s$ and control input $u$ defined as follows: 
\begin{equation}
    s=\begin{bmatrix}
        p_x \\ p_y \\ \theta
    \end{bmatrix} \in \mathcal{S}, \ u=\begin{bmatrix}
        v \\ \omega
    \end{bmatrix} \in \mathcal{U}, \ 
    \label{eqn::robotstate}
\end{equation}
where $p_x$ and $p_y$ denote the $x$-axis (longitudinal) and $y$-axis (lateral) position in the global coordinate system, respectively; $\theta$ denotes the heading angle of the robot; $v$ denotes the velocity of the robot; and $\omega$ denotes the changing rate of the heading angle of the robot. 
    
The simplified unicycle model in discrete time for the mobile robot is given as follows: 
    \begin{equation}
        s(k+1)=f(s(k),u(k),\Delta t),
        \label{eqn::ctrl system}
    \end{equation}
where $k$ denotes the time step; and $\Delta t$ denotes the discrete time step. $f(s(k),u(k),\Delta t)$ is defined as follows: 
    \begin{equation}
    \begin{aligned}
        \begin{bmatrix} p_{x,k+1} \\ p_{y,k+1} \\ \theta_{k+1} \end{bmatrix} &= \begin{bmatrix} p_{x,k} + v_k \Delta t \cos(\theta_k) \\ p_{y,k} + v_k \Delta t \sin(\theta_k) \\ \theta_k + \omega_k \Delta t \end{bmatrix}.
        \label{eqn::ctrl system detail}
    \end{aligned}
    \end{equation}

We make the following assumption for this problem: 

\begin{assumption}
    When an obstacle's center is not visible in the robot's FoV, it is treated as an occluded obstacle.
\end{assumption}
\begin{assumption}
The velocities of dynamic obstacles are bounded, and their maximum velocities are known.
\end{assumption}

  The objective of this work is to develop a motion planning strategy that generates a set of trajectories, enabling safe and consistent navigation in occluded, obstacle-dense environments. The proposed strategy must satisfy the following criteria when navigating in occluded, obstacle-dense environments: 
    
    \textit{\textbf{Computational efficiency}: }
    Ensure real-time parallel trajectory generation.
    
    \textit{\textbf{Safety guarantee}: }
    Guarantee collision-free navigation. 

    \textit{\textbf{Motion consistency}: }
Ensure consistent trajectory generation, avoiding significant velocity fluctuations and maintaining smooth motion. 

\section{Methodology}\label{Methodology}
In this section, we propose an occlusion-aware CMPC strategy for mobile robots navigating in occluded, obstacle-dense environments. 

\subsection{Risk Regions Modeling}
To address the uncertainty of the states of occluded obstacles, we dynamically model risk regions to represent potential locations of occluded obstacles. 

\subsubsection{Definition of Occluded Regions}\label{section:occluded regions}
Occluded regions are defined as areas blocked by obstacles within the robot's FoV. For each occluded region, in the robot's body frame with its center as the origin, the occluded region is defined as the set of points that satisfy the following conditions: 
\begin{itemize}
    \item The points lie between two tangent lines of the obstacle originating from the origin.
    \item The points are located behind the obstacle.
\end{itemize}

Let $m_1$ and $m_2$ denote the slopes of the two tangent lines that form the boundaries of the occluded region, $x_{\text{rel}}$ and $y_{\text{rel}}$ denote the position of the obstacle in the robot's body frame, and $r_{\text{obs}}$ denotes the radius of the obstacle. The expressions for the two tangent lines are defined as follows:
\begin{subequations}
    \begin{align}
        y &=m_1x,\\
        y &=m_2x,
    \end{align}    
\end{subequations}
where 
\begin{subequations}
    \begin{align}
        m_1 &=\frac{x_{\text{rel}} \cdot y_{\text{rel}}+r_{\text{obs}}\cdot \sqrt{x_{\text{rel}}^2+y_{\text{rel}}^2-r_{\text{obs}}^2}}{x_{\text{rel}}^2-r_{\text{obs}}^2}, \\
        m_2 &=\frac{x_{\text{rel}} \cdot y_{\text{rel}}-r_{\text{obs}}\cdot \sqrt{x_{\text{rel}}^2+y_{\text{rel}}^2-r_{\text{obs}}^2}}{x_{\text{rel}}^2-r_{\text{obs}}^2}.  
    \end{align}    
\end{subequations}

The occluded region $\mathcal{C}$ is represented by the set of points $\boldsymbol{x} = \begin{bmatrix} x \\ y \end{bmatrix}$ in the 2D plane that satisfy the following condition: 
\begin{equation}
    \mathcal{C}=\{\boldsymbol{x}\in \mathbb{R}^2 | \boldsymbol{A} \boldsymbol{x}<0\}, \label{eqn:occluded region}
\end{equation}
where $\boldsymbol{A}=\begin{bmatrix}-m_1 & 1 \\ m_2 & -1 \\ -1 & 0\end{bmatrix}$ denotes the coefficient matrix that represents the boundaries of the occluded region.

\begin{figure}[tbp]
\centering
\includegraphics[width=0.517\columnwidth]{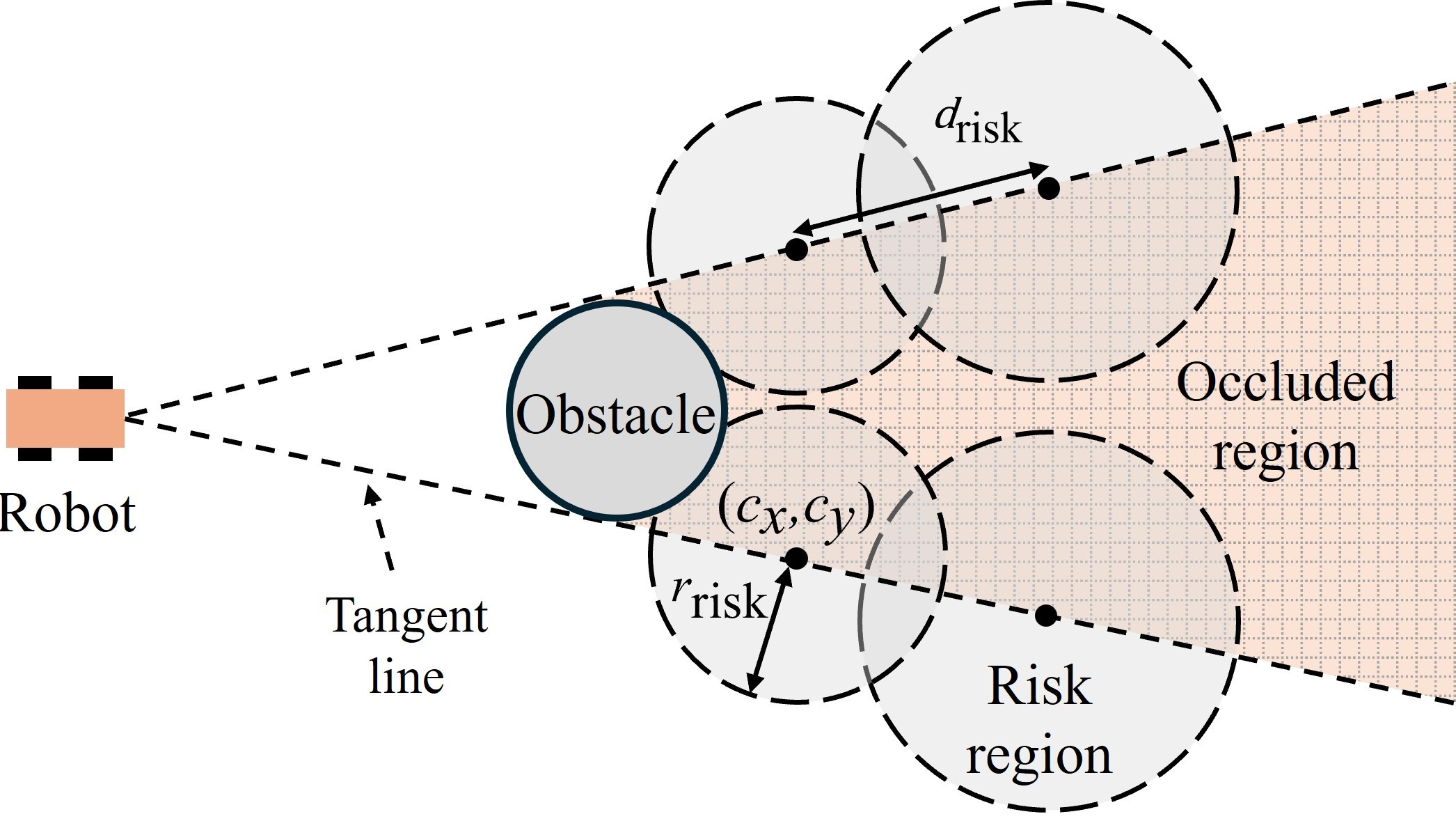}
    \caption{Modeling of occluded regions and risk regions. Occluded regions are bounded by two tangent lines. Risk regions are located along two tangent lines, each with center $(c_x,c_y)$ and radius $r_{\text{risk}}$. }
    \label{fig:riskregion}
    \vspace{-1em}
\end{figure}

\subsubsection{Definition of Risk Regions}
Based on the defined occluded regions, we further delineate several risk regions that affect the robot's motion. Typically, obstacles at the edge of the occluded regions pose the greatest safety threat to the robot. Therefore, these risk regions are defined along the two tangent lines that form the occluded region described in Section~\ref{section:occluded regions}. Specifically, to provide an effective geometric approximation of the polygonal risk cones along the tangent lines, we model two circular risk regions on each tangent line, as shown in Fig.~\ref{fig:riskregion}. For the $i$-th circular risk region, the center $(c_x^{(i)},c_y^{(i)})$ is defined as follows: 
\begin{equation}
\label{eqn:centers}
\begin{aligned}
\begin{bmatrix}c_x^{(i)}\\c_y^{(i)}\end{bmatrix}
=&\left(i d_{\text{risk}}+\sqrt{d_{\text{obs}}^2-r_{\text{obs}}^2}\right)
\cdot \boldsymbol{R}\,
\frac{1}{\sqrt{1+m_j^2}}
\begin{bmatrix}1\\ m_j\end{bmatrix}\\
&+\begin{bmatrix}p_x\\p_y\end{bmatrix}, \quad 
j \in \mathcal{I}_1^2,\ i \in \mathcal{I}_0^{N_{\text{risk}}-1},
\end{aligned}
\end{equation}
where $m_j$ denotes the slope of the $j$-th tangent line of the occluded region; $d_{\text{risk}}$ denotes the distance between consecutive risk regions; $d_{\text{obs}}$ denotes the distance between the robot and the obstacle; $N_{\text{risk}}$ denotes the total number of risk regions in each configuration; and $\mathcal{I}_{a}^{b}$ denotes a sequence of integers from $a$ to $b$. Since the tangent lines are defined in the robot's body frame, we transform them to the global frame via a rotation matrix $\boldsymbol{R}$.

The radius of the $i$-th risk region $r_{\text{risk}}^{(i)}$ is determined as follows: 
\begin{equation}
    r_{\text{risk}}^{(i)}=\frac{\left\|\,(c_x^{(i)},c_y^{(i)})-(p_x,p_y)\,\right\|_2}{v+\sigma}\cdot v_{\text{obs,max}}+r_{\text{obs}}, 
    \label{eqn::risk_radius}
\end{equation}
where $\sigma \in\mathbb{R}^+$ is a small regularization constant (e.g., $\sigma =10^{-4}$) that ensures numerical stability; $v$ denotes the velocity of the robot; and $v_{\text{obs,max}}$ denotes the assumed maximum velocity of the occluded obstacle. 

In this study, the conservatism in trajectory planning can be adjusted via different assumptions about the maximum velocity $v_{\text{obs,max}}$ to generate a sequence of $N_z$ risk region configurations.

\begin{remark}
By utilizing the maximum possible obstacle velocity $v_\text{obs,max}$ in~\eqref{eqn::risk_radius}, our circular risk region model provides a conservative over-approximation that encompasses potential positions an occluded obstacle may reach during the planning horizon. For non-circular obstacles, we use their minimum bounding circles to define the risk regions, ensuring the generality of our formulation. 
\end{remark}

\subsection{Occlusion-Aware CMPC} 
This section introduces an occlusion-aware CMPC strategy with multiple branches emanating from a common consensus segment. Each branch is configured with a distinct risk region configuration within the entire set of $N_z$ risk region configurations, enabling the exploration of diverse scenarios while maintaining a consensus initial trajectory segment.

To complete the navigation task and enhance motion consistency, we define a cost function $\mathcal{J}$ over a planning horizon with $N$ time steps to address these requirements: 
\begin{equation}
    \begin{aligned}
        \mathcal{J}=&\sum_{k=0}^{N-1}[ w_{\text{acc}}C_{\text{acc}}(s(k))+w_{\text{vel}}C_{\text{vel}}(s(k))]\\&+w_{\text{guide}}C_{\text{guide}}(s(N-1)),
        \label{eqn::cost}
    \end{aligned}
    \end{equation}
    where $C_\text{acc}(\cdot)$ regulates the robot's acceleration; $C_\text{vel}(\cdot)$ represents the velocity tracking cost, measured by the velocity error between the robot and reference velocity; and $C_\text{guide}(\cdot)$ represents the deviation from the guidance points. $w_{\text{acc}},\ w_{\text{vel}},\ w_{\text{guide}} \in \mathbb{R}^+$ are the corresponding weights for the cost components. 
    
    \begin{remark}
    A set of guidance points is generated using a Visual-PRM approach~\cite{10160379}. It produces optimal guidance paths for the receding horizon based on the global task, considering all visible obstacles. Specifically, guidance points provide the endpoints of each horizon, so the cost $C_\text{guide}(\cdot)$ is only calculated at the terminal time step $k=N-1$.  
    \end{remark}
    
The motion planning problem is reformulated as a nonlinear optimization problem that minimizes the cost function $\mathcal{J}$ in~\eqref{eqn::cost}, subject to the robot's kinematic, collision avoidance, and risk boundary constraints as follows: 
    \begin{subequations}
        \begin{align}
        \displaystyle\operatorname*{\text{min}}_{s,u}~~~ &\mathcal{J} \label{eqn:cost objective2}\\
        \text{s.t.}\quad 
        &H_{\text{kin}}(s)=0, \label{eqn:constraint0_1} \\
        &G_{\text{obs}}(s) \leq 0,\\
        &G_{\text{risk}}(s)\leq 0,\\
        &s(k)=\Tilde{s}(k),\quad \forall k\in\mathcal{I}_0^{N_c-1},\label{eqn:consensus_}
        \end{align}
    \end{subequations}
where
\begin{subequations}
    \begin{align}
        &H_{\text{kin}}(s)=s(k+1)-f(s(k),u(k),\Delta t),\\ & \forall k \in \mathcal{I}_0^{N-1},\label{eqn:Hkin}\\
        &G_{\text{obs}}(s) = r_{\text{obs}}^2 - \|(p_x,p_y) - (x_{\text{obs}},y_{{\text{obs}}
        })\|_2^2,\ \notag\\ &\forall \{(x_{\text{obs}},y_{\text{obs}}),r_{\text{obs}}\}\in  \{\mathcal{B}^{(j)}\} _{j=0}^{N_{\text{obs}}-1},\ \forall s \in \mathcal{S},\label{eqn:Gobs}\\
        &G_{\text{risk}}(s) = (r_{\text{risk}}^{(i)})^2 - \|(p_x,p_y) - (c_x^{(i)},c_y^{(i)})\|_2^2,\ \notag\\ &\forall \{(c_x^{(i)},c_y^{(i)}),r_{\text{risk}}^{(i)}\} \in \{ \mathcal{R}^{(i)}_z\}_{i=0}^{N_{\text{risk}}-1},\ \forall s \in \mathcal{S}. \label{eqn:Grisk}
    \end{align}
\end{subequations}

Here, $H_{\text{kin}}(\cdot)$ in~\eqref{eqn:Hkin} enforces kinematic constraints derived from the robot's kinematic model defined in~\eqref{eqn::ctrl system detail}; $G_{\text{obs}}(\cdot)$ in~\eqref{eqn:Gobs} ensures a minimum safety distance $r_{\text{obs}}$ between the robot and obstacle centers to avoid collision; $\mathcal{B}^{(j)}$ represents the state of the $j$-th obstacle, with the position $(x_{\text{obs}}, y_{\text{obs}})$ and the radius of the obstacle $r_{\text{obs}}$; $N_{\text{obs}}$ denotes the number of obstacles considered; $G_{\text{risk}}(\cdot)$ in~\eqref{eqn:Grisk} maintains a risk boundary $r_{\text{risk}}^{(i)}$ around risk regions; $\mathcal{R}^{(i)}_{z}$ represents the states of the $i$-th risk region in the $z$-th configuration, with center and radius derived from~\eqref{eqn:centers} and~\eqref{eqn::risk_radius}; $N_z$ denotes the total number of risk region configurations; \eqref{eqn:consensus_} represents the consensus constraint, with consensus variable $\Tilde{s}$; and $N_c$ denotes the number of time steps in the consensus segment.

\begin{remark}
   The consensus variable $\Tilde{s}$ is identical across all branches in the consensus segment, ensured by~\eqref{eqn:consensus_}. 
   This approach ensures that the trajectory inside the consensus segment is unified while allowing for divergent trajectories in subsequent segments, thereby enabling dynamic scenario exploration while ensuring motion consistency. 
\end{remark}

\subsection{ADMM-based Optimization}
To facilitate high computational efficiency, we decompose the motion planning problem \eqref{eqn:cost objective2}-\eqref{eqn:consensus_} into several low-dimensional sub-problems using the Jacobi-Proximal ADMM scheme. This method allows for independent parallel optimization of each sub-problem, with the consensus constraint ensuring synchronization across them. 

While ADMM is effective in handling separable problems, the enforcement of complex constraints typically requires more sophisticated mechanisms. To address this, the augmented Lagrangian method (ALM) is employed.
It integrates primal and dual variables to enforce constraints more strictly. For the inequality constraints $G_{\text{obs}}(\cdot)$ and $G_{\text{risk}}(\cdot)$, we address them by incorporating squared penalty terms~\cite{toussaint2014novelaugmentedlagrangianapproach}. For each sub-problem with the $z$-th risk region configuration, the augmented Lagrangian function is defined as follows: 
\begin{equation}
\begin{aligned}
    &L_z(s_z,\Tilde{s},\lambda_{\text{obs},z},\lambda_{\text{risk},z},\lambda_{\text{kin},z},\lambda_{\text{cons},z},\rho_{\text{obs}},\rho_{\text{risk}},\rho_{\text{kin}},\rho_{\text{cons}})=\\
    & \  \mathcal{J} 
    +\lambda_{\text{obs},z}^{\intercal}G_{\text{obs},z}(s_z)+\rho_{\text
    {obs}} \Vert \mathcal{I}_+(G_{\text{obs},z}(s_z))\cdot G_{\text{obs},z}(s_z) \Vert^2\\
    &+\lambda_{\text{risk},z}^{\intercal}G_{\text{risk},z}(s_z)+\rho_{\text
    {risk}} \Vert \mathcal{I}_+(G_{\text{risk},z}(s_z))\cdot G_{\text{risk},z}(s_z) \Vert^2\\ 
    &+\lambda_{\text{kin},z}^{\intercal}H_{\text{kin},z}(s_z)+\rho_{\text{kin}} \Vert H_{\text{kin},z}(s_z) \Vert^2\\
    &+\lambda_{\text{cons},z}^{\intercal}(s_z-\Tilde{s}) +\rho_{\text{cons}} \Vert s_z-\Tilde{s} \Vert ^2, \label{eqn:ALM}
    \end{aligned}
    \end{equation}
where $s_z$ is the robot's state in the sub-problem with the $z$-th risk region configuration. $G_{\text{obs},z}(\cdot)$ and $G_{\text{risk},z}(\cdot)$ represent inequality constraints for obstacle avoidance~\eqref{eqn:Gobs} and risk boundary~\eqref{eqn:Grisk}, respectively. $H_{\text{kin},z}(\cdot)$ represents the kinematic equality constraint~\eqref{eqn:Hkin}.  $\lambda_{\text{obs},z}$, $\lambda_{\text{risk},z}$, $\lambda_{\text{kin},z}$, $\lambda_{\text{cons},z}$ are the corresponding dual variables;  $\rho_{\text{obs}}$, $\rho_{\text{risk}}$, $\rho_{\text{kin}}$, $\rho_{\text{cons}}$ are the corresponding penalty coefficients. Specifically, $\rho_{\text{cons}}$ is the penalty parameter for the proximal term of the consensus constraint. $\mathcal{I}_+(F)$ is an indicator function defined as follows: 
\begin{equation}
    \mathcal{I}_+(F) = \begin{cases}
    1, & \text{if } F
    >0, \\
    0, & \text{otherwise}.
\end{cases}
\end{equation}

\setlength{\textfloatsep}{ 1.\baselineskip plus  0.2\baselineskip minus  0.4\baselineskip}
\begin{algorithm}[t]
\caption{Occlusion-Aware CMPC}\label{alg:opt_process}
\begin{algorithmic}[1]
\State \textbf{While} task not done \textbf{do}
\State \hspace{1em} Obtain the states of the robot $s$, visible obstacles \Statex \hspace{1em} $\{ \mathcal{B}^{(j)}\} _{j=0}^{N_{\text{obs}}-1}$ and $N_z$ configurations of risk regions \Statex \hspace{1em} $\{ \mathcal{R}^{(i)}_{z}\} _{z=0}^{{N_z}-1},\ i\in\mathcal{I}_0^{N_{\text{risk}}-1}$ via~\eqref{eqn:occluded region}-\eqref{eqn::risk_radius}; 
\Statex \hspace{1em} \textbf{ADMM-based Optimization}
\State \hspace{2em} Reformulate as $N_z$ sub-problems, each with the \Statex \hspace{2em} $z$-th risk region configuration via~\eqref{eqn:ALM}; 
\State \hspace{2em} \textbf{For} \(\iota \gets 0\) \textbf{to} \(\iota_{\text{max}}\) \textbf{do}
\Statex \hspace{3em} \textbf{Parallel Optimization} of all sub-problems
\State \hspace{4em} Solve for $s_z$ via~\eqref{eqn:unconstrainted}; 
\State \hspace{4em} Update dual variables $\lambda_{\text{obs},z},\lambda_{\text{risk},z},\lambda_{\text{kin},z}$ \Statex \hspace{4em} via \eqref{eqn:constrained1}-\eqref{eqn:constrained3}; 
\Statex \hspace{3em} \textbf{End Parallel Optimization}
\State \hspace{3em} Update consensus variable $\Tilde{s}$ via~\eqref{eqn:consensus update}; 
\State \hspace{3em} Update ADMM dual variable $\lambda_{\text{cons},z}$ via~\eqref{eqn:ADMMupdate};
\State \hspace{3em} \textbf{Break if} the termination criteria~\eqref{eqn:ALMdualtol}-\eqref{eqn:ADMMdualtol} \State \hspace{3em} are met;
\State \hspace{2em} \textbf{End For} 
\State \hspace{1em} Apply the first control input from the optimized  \Statex \hspace{1em} consensus segment to the robot via \eqref{eqn::ctrl system detail}; 
\State \textbf{End While}
\end{algorithmic}
\end{algorithm}

Note that the sub-problems are independent with respect to different risk region configurations, except for the consensus constraint~\eqref{eqn:consensus_}. Meanwhile, the updating process of consensus variable $\Tilde{s}$ depends on the results of all sub-problems. Therefore, the optimization problem can be decomposed into $N_z$ independent sub-problems. The consensus variable $\Tilde{s}$ is updated once all sub-problems are optimized. 

For each sub-problem of the CMPC with the $z$-th risk region configuration, the optimization process is as follows: 
\begin{subequations}\label{eqn:subupdate}
\vspace{-1.5em}
    \begin{align}    
s_z^{\iota+1}=&\argmin{s_z,u_z}L_z(\cdot), \label{eqn:unconstrainted} \\
        \lambda_{\text{obs},z}^{\iota+1}=&\lambda_{\text{obs},z}^{\iota}+2\rho_{\text{obs}}G_{\text{obs},z}(s_z^{\iota+1}),\label{eqn:constrained1}\\
        \lambda_{\text{risk},z}^{\iota+1}=&\lambda_{\text{risk},z}^{\iota}+2\rho_{\text{risk}}G_{\text{risk},z}(s_z^{\iota+1}),\label{eqn:constrained2}\\
        \lambda_{\text{kin},z}^{\iota+1}=&\lambda_{\text{kin},z}^{\iota}+2\rho_{\text{kin}}H_{\text{kin},z}(s_z^{\iota+1}),\label{eqn:constrained3}
    \end{align}
\end{subequations}
where $\iota$ denotes the index of iterations. 

For each sub-problem,~\eqref{eqn:unconstrainted} is optimized using Newton's method~\cite{gill2012primal}. The dual variables corresponding to the constraints are updated in~\eqref{eqn:constrained1}-\eqref{eqn:constrained3}. The optimizations of all sub-problems are performed in parallel. 

Once the optimizations in~\eqref{eqn:unconstrainted}-\eqref{eqn:constrained3} of all sub-problems are complete, we update the consensus variable $\Tilde{s}$ and ADMM dual variable $\lambda_{\text{cons},z}$ as follows: 
\begin{subequations}
    \begin{flalign}
        \Tilde{s}^{\iota+1}=&\frac{1}{N_z}\sum_{z=0}^{N_z-1} s_z^{\iota+1}, \label{eqn:consensus update} \\
        \lambda_{\text{cons},z}^{\iota+1}=&\lambda_{\text{cons},z}^{\iota}+2\rho_{\text{cons}}(s_z^{\iota+1}-\Tilde{s}^{\iota+1}), \label{eqn:ADMMupdate}
    \end{flalign}
\end{subequations}
where consensus variable $\Tilde{s}$ is updated in~\eqref{eqn:consensus update} by averaging the results from each sub-problem within the consensus segment. The ADMM dual variable $\lambda_{\text{cons},z}$ is updated in~\eqref{eqn:ADMMupdate} for each sub-problem.

The optimization process terminates once the constraints are satisfied in each sub-problem and the consensus segment is constructed. The termination criteria are defined as follows: 
    \begin{subequations}
    \vspace{-1em}
        \begin{align}
            \Vert \nabla L_z(\cdot) \Vert \leq &\epsilon^{\text{dual}},\quad\forall z\in\mathcal{I}_0^{N_z-1},\label{eqn:ALMdualtol}\\
            \Vert s_z^{\iota+1}-\Tilde{s}^{\iota+1} \Vert \leq &\xi^{\text{pri}},\quad\forall z\in\mathcal{I}_0^{N_z-1},\label{eqn:ADMMpritol}\\
            \Vert \nabla \Tilde{s}\Vert \leq &\xi^{\text{dual}}, \label{eqn:ADMMdualtol}
        \end{align}
    \end{subequations}
where $\Vert \nabla L_z(\cdot) \Vert$ denotes the norm of the gradient of augmented Lagrangian from each sub-problem; $\Vert \nabla \Tilde{s}\Vert$ denotes the norm of the gradient of consensus variable $\Tilde{s}$; 
$\epsilon^{\text{dual}}$ denotes the dual residual threshold of ALM; $\xi^{\text{pri}}$ and $\xi^{\text{dual}}$ denote the primal and dual residual thresholds of ADMM, respectively. Based on the practical experimental performance, the values of $\epsilon^{\text{dual}}$, $\xi^{\text{pri}}$ and $\xi^{\text{dual}}$  are set to $0.15$, $0.1$ and $0.1$, respectively. The penalty coefficients $\rho_{\text{obs}}$, $\rho_{\text{risk}}$, $\rho_{\text{kin}}$, $\rho_{\text{cons}}$ are set to $1.0$; and all dual variables are initialized to $0.0$.

The optimization process continues until the termination criteria~\eqref{eqn:ALMdualtol}-\eqref{eqn:ADMMdualtol} are met for all sub-problems and the consensus segment, or the maximum iteration $\iota_{\text{max}} = 300$ is reached. A detailed description of the CMPC optimization process is provided in \textbf{Algorithm}~\ref{alg:opt_process}. 

\setlength{\textfloatsep}{ 1.7\baselineskip plus  0.2\baselineskip minus  0.4\baselineskip}
 \begin{figure*}[tb]
	 	\centering  
        \subfigcapskip = -0mm
        \subfigure[Time instant $t = 8\,\text{s}$. \label{fig:sim_exp1}]{
		\includegraphics[width=0.23\textwidth]{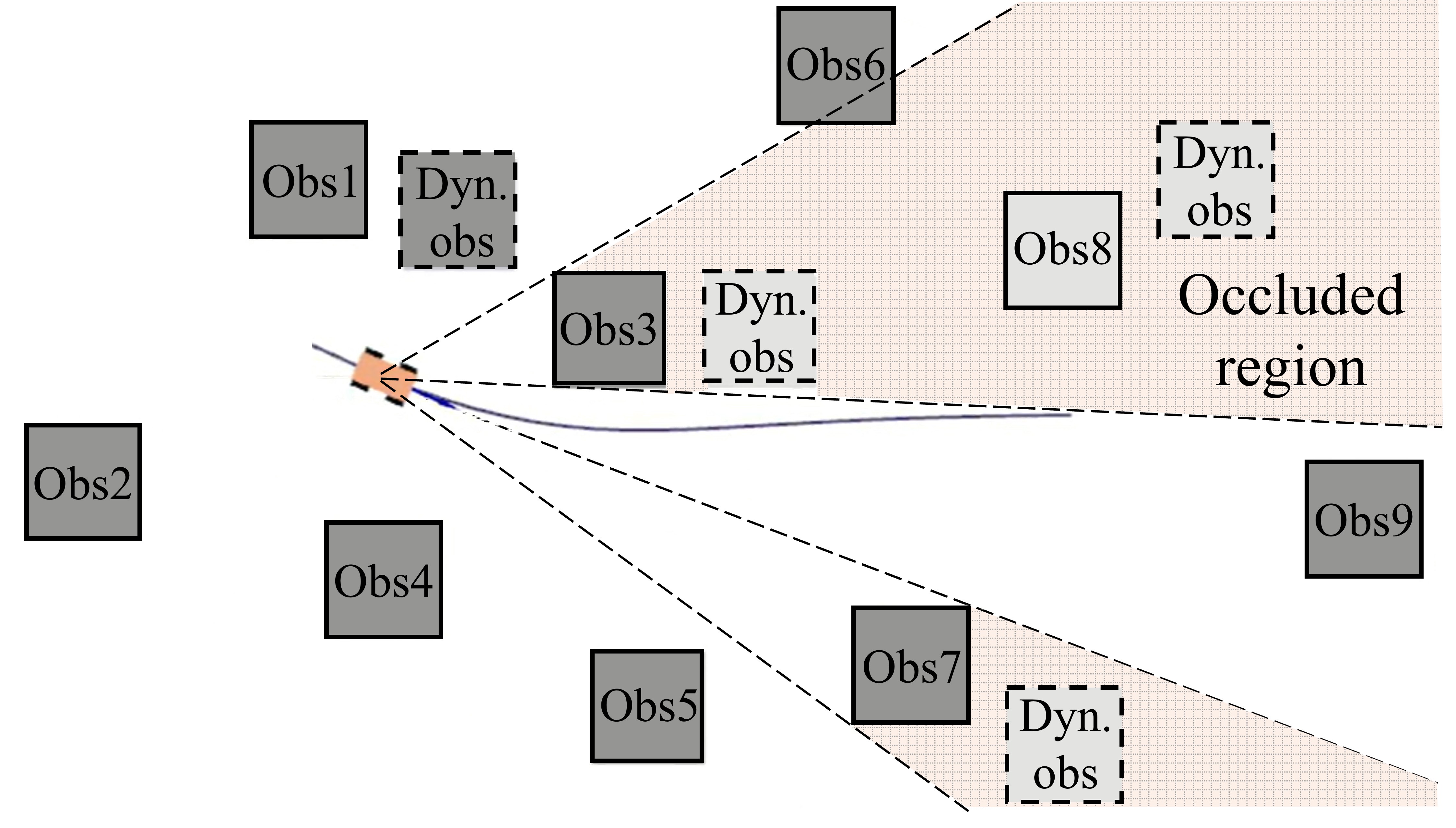}}
               \subfigure[Time instant $t = 10\,\text{s}$.\label{fig:sim_exp2} ]{
			
			\includegraphics[width=0.23\textwidth]{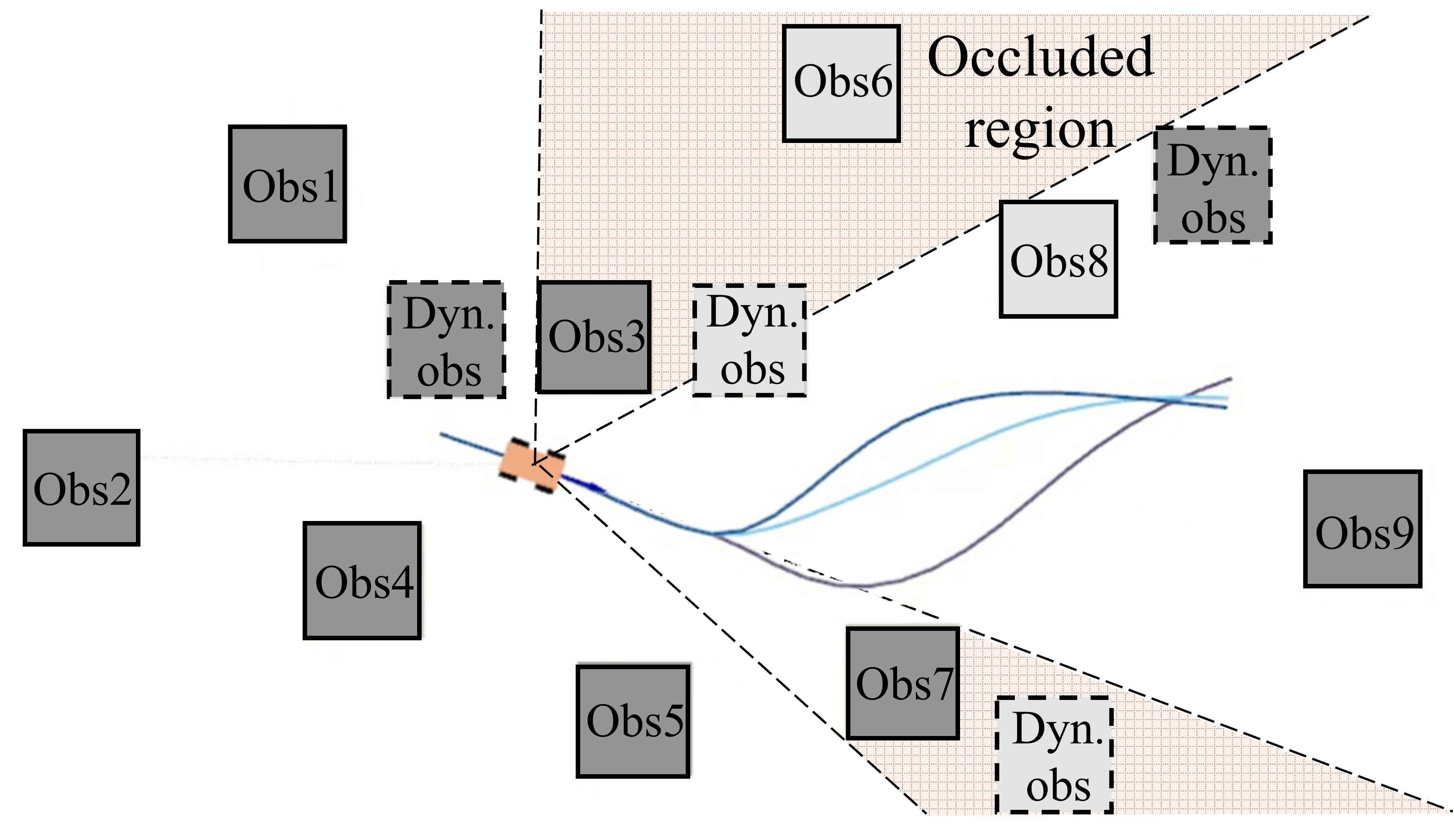}}
        \subfigure[Time instant $t = 13\,\text{s}$.\label{fig:sim_exp3} ]{
        \includegraphics[width=0.23\textwidth]{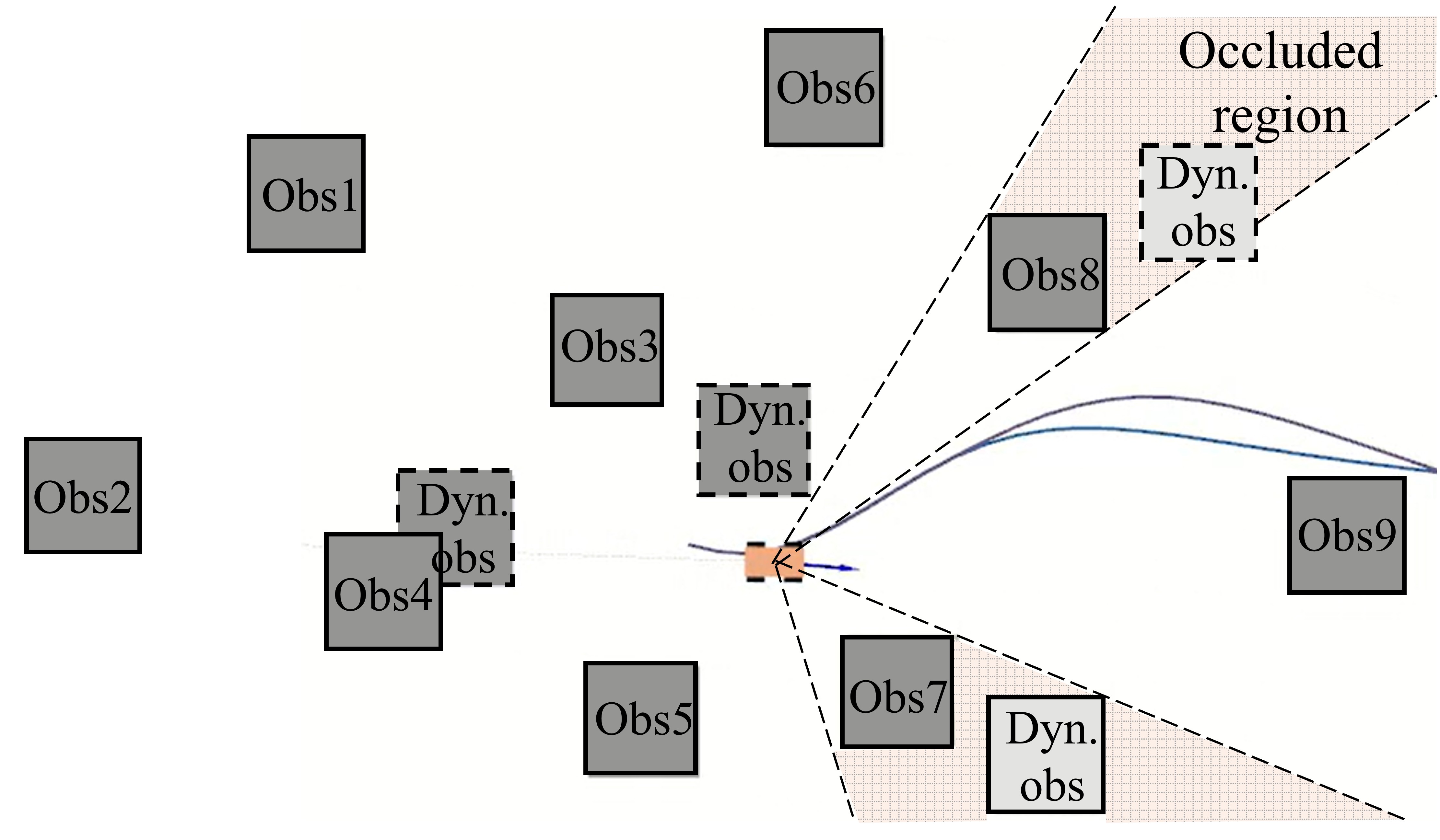}
            }
            \subfigure[Time instant $t = 17\,\text{s}$.\label{fig:sim_exp4} ]{\includegraphics[width=0.23\textwidth]{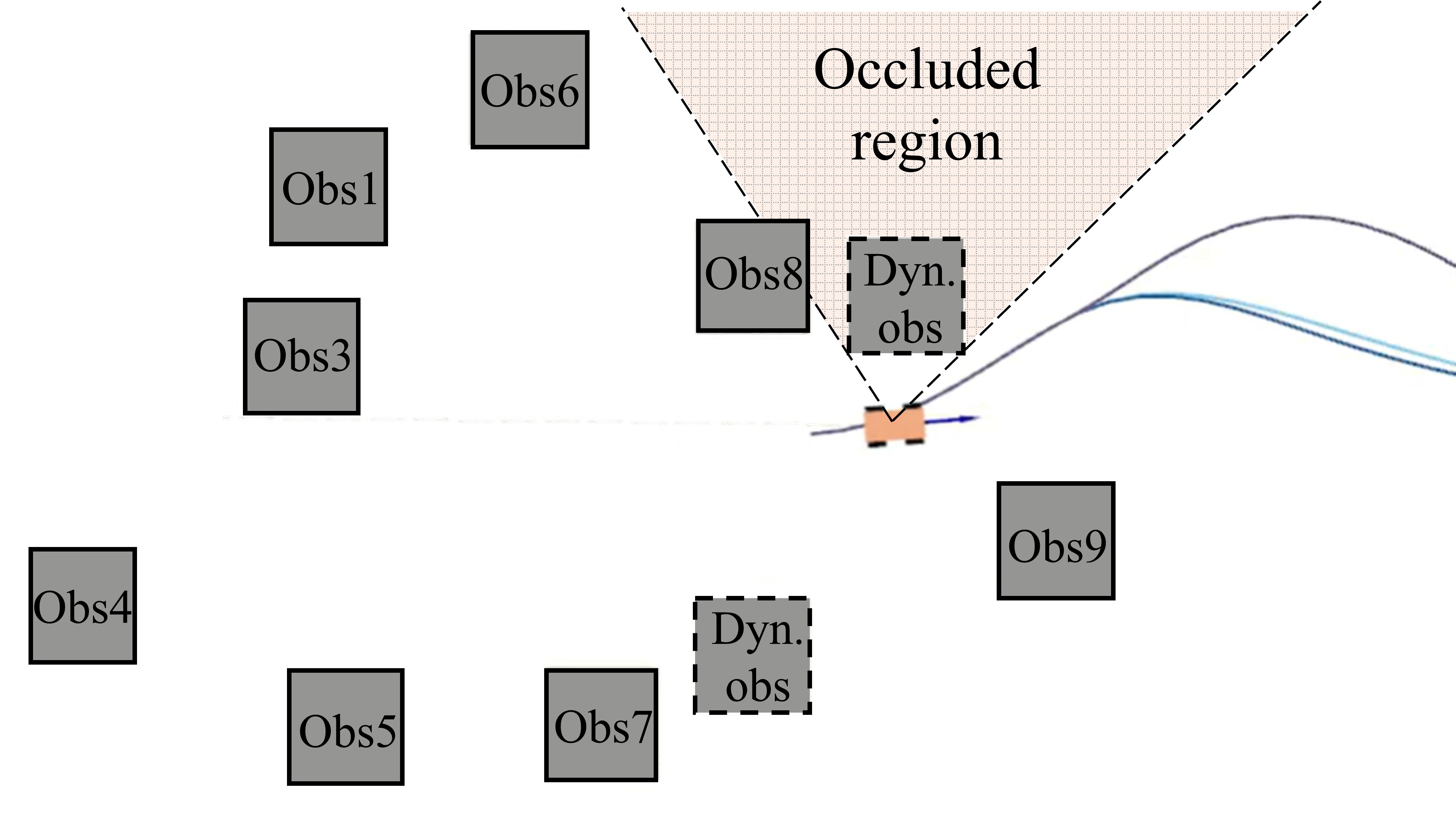}
            }
            \vspace{-1mm}
	 	\caption{Simulation snapshots, where the robot navigates in an occluded, obstacle-dense environment. The robot, shown as an orange rectangle with black wheels, optimizes three trajectories in parallel. Dark-gray boxes represent visible obstacles, and light-gray boxes represent occluded obstacles. The arrow indicates the current velocity vector of the robot. Curves in different colors represent trajectories with different risk region configurations. }		
	 	\label{fig:sim_exp}
\end{figure*}	

\begin{table*}[tp]
    \centering
    \scriptsize
    \caption{Quantitative Results Comparison Among Different Approaches}
    \label{tab:comparasion_results}
    
    \begin{tabular}[c]{ccccc}
        \hline       
        \textbf{Algorithm} & \textbf{Collision} &\textbf{Max Lat. Vel. Variance } ($\text{m/s}$) & \textbf{Peak Lat. Acc. }  
        ($\text{m/s}^2$)& \textbf{Avg. Solving Time} ($\text{ms}$)\\
        \hline
        Control-Tree &\text{YES} & $3.34$ &  $7.34$ & $45.03$\\
        \hline
       Single hypothesis MPC without risk region  & \text{YES}  & $2.27$ & $3.67$ & $19.84$ \\
        \hline
       Single hypothesis MPC with risk region  & \text{NO}  & $3.04$ & $7.56$ & $21.15$  \\
        \hline
        CMPC-0 & \text{NO} & $2.55$ & $7.01$ & $38.01$ \\
        \hline
        \textbf{CMPC}  & \textbf{NO}  & \textbf{$\mathbf{1.88}$} & \textbf{$\mathbf{3.65}$} & \textbf{$\mathbf{40.78}$}  \\
        \hline        
    \end{tabular}
    \vspace{-0.5em}
\end{table*}

\begin{table}
    \centering
    \scriptsize
    \caption{Ablation Study Among Different Consensus Segment Length}
    \label{tab:comparasion_consensus}
    \resizebox{\columnwidth}{!}{
    \begin{tabular}[c]{cccc}
        \hline
\makecell{\textbf{Consensus Segment}\\\textbf{Length $T_c$} ($\text{s}$)} &\makecell{\textbf{Max Lat.} \\ \textbf{Vel. Variance } ($\text{m/s}$)} & \makecell{\textbf{Peak Lat.} \\ \textbf{ Acc.}  
        ($\text{m/s}^2$)}& \makecell{\textbf{Avg.} \\\textbf{Solving Time} ($\text{ms}$)}\\
        \hline
        0 & $2.55$ &  $7.01$ & $38.01$\\
        \hline
       1  & $2.34$ & $4.74$ & $39.32$ \\
        \hline
       2    & $1.88$ & $3.65$ & $40.78$  \\
        \hline
        5   & $2.22$ & $5.28$ & $47.18$  \\
        \hline     
    \end{tabular}}
    \vspace{-0.5em}
\end{table}

\begin{remark}
The computational complexity of the algorithm for each iteration can be specified as:
\begin{itemize}
    \item The state variable update exhibits $\mathcal{O}(N)$ complexity. 
    \item Averaging operations over the consensus segment scale as $\mathcal{O}(N_c \cdot m)$. 
    \item The obstacle-related and risk region-related dual variable updates~\eqref{eqn:constrained1} and ~\eqref{eqn:constrained2} scale as $\mathcal{O}(N \cdot  N_\text{obs})$ and $\mathcal{O}(N \cdot N_\text{risk})$, respectively. Dual variable updates~\eqref{eqn:constrained3} and ~\eqref{eqn:ADMMupdate} scale as $\mathcal{O}(N)$ and $\mathcal{O}(N_c)$, respectively.
\end{itemize}

Given that the number of time steps $N$ in the planning horizon and the number of time steps $N_c$ in the consensus segment are fixed, and the number of risk regions $N_\text{risk}$ is bounded, the computational complexity per ADMM iteration scales with the number of obstacles as $\mathcal{O}(N_\text{obs})$. 
\end{remark}


\begin{remark}
In the Jacobi-Proximal ADMM-based optimization scheme, each sub-problem optimizes a single trajectory branch with its specific risk region configuration.  
In each iteration, the optimization step \eqref{eqn:unconstrainted} is solved in parallel for each risk region configuration $z$, yielding the optimal states $s_z$ for each trajectory branch. This parallelization is key to achieving real-time performance. The solutions of all branches are then aggregated through the consensus update~\eqref{eqn:consensus update}–\eqref{eqn:ADMMupdate}, which enforces that all branches share a common consensus segment, thus ensuring motion consistency. Additionally, the proximal term $\rho_{\text{cons}} \| s_z - \Tilde{s} \|^2$ in the augmented Lagrangian \eqref{eqn:ALM} stabilizes the convergence of this parallel process, and this ensures numerical robustness.
\end{remark}

\section{Experiments}\label{experiments}

\subsection{Simulation}\label{section:exp_in_sim}
\subsubsection{Simulation Setup}
The simulations are conducted using C++ and ROS1 on an Ubuntu 20.04 system, equipped with an Intel Ultra7 155H CPU (16 cores @ 1.40 GHz) and 16 GB of RAM. The simulation environment utilizes Gazebo 9 for dynamic simulation and RVIZ for visualization of the robot's trajectories and motion. The robot's dimensions are $800\,\text{mm} \times 400\,\text{mm}$. Blocks sized $1500 \, \text{mm} \times 1500 \,\text{mm}$ serve as both static and dynamic obstacles. 

The robot is tasked with navigating through an occluded, obstacle-dense area containing dynamic and static obstacles. Some dynamic obstacles are occluded from the robot's FoV, such as the light-gray boxes inside the occluded region shown in Fig.~\ref{fig:sim_exp1} and Fig.~\ref{fig:sim_exp2}. Dynamic obstacles initiate movement at a velocity of $0.6\,\text{m/s}$ to $1.0\,\text{m/s}$ along the lateral direction when their distance to the robot is within $2\, \text{m}$. The planning horizon is $T_h=6\,\text{s}$ with $4$ steps per second, resulting in $N = 24$ time steps; and the consensus segment length is $T_{c}=2\,\text{s}$, resulting in $N_c=8$ time steps. The reference velocity of the robot is set to $1.8\,\text{m/s}$. The weighting matrices are set to $w_{\text{guide}} = 3.5$, $w_{\text{vel}} = 5.0$, $w_{\text{acc}} = 1.8$. 

We set three branches of trajectories for the robot, each accounting for different risk region configurations. The first trajectory neglects all risk regions to maximize task efficiency; the second considers risk regions with a maximum possible velocity of the obstacles $v_{\text{obs,max}}=0.5\,\text{m/s} $ in~\eqref{eqn::risk_radius} to tackle a relatively common scenario, and the third considers risk regions with $v_{\text{obs,max}}=1\,\text{m/s}$ to maximize safety awareness. Each trajectory accounts for at most two occluded regions of the nearest obstacles and their corresponding eight risk regions (four risk regions for each occluded region). 

\begin{figure}[tbp] 
    \centering 
\includegraphics[width=0.86\columnwidth]{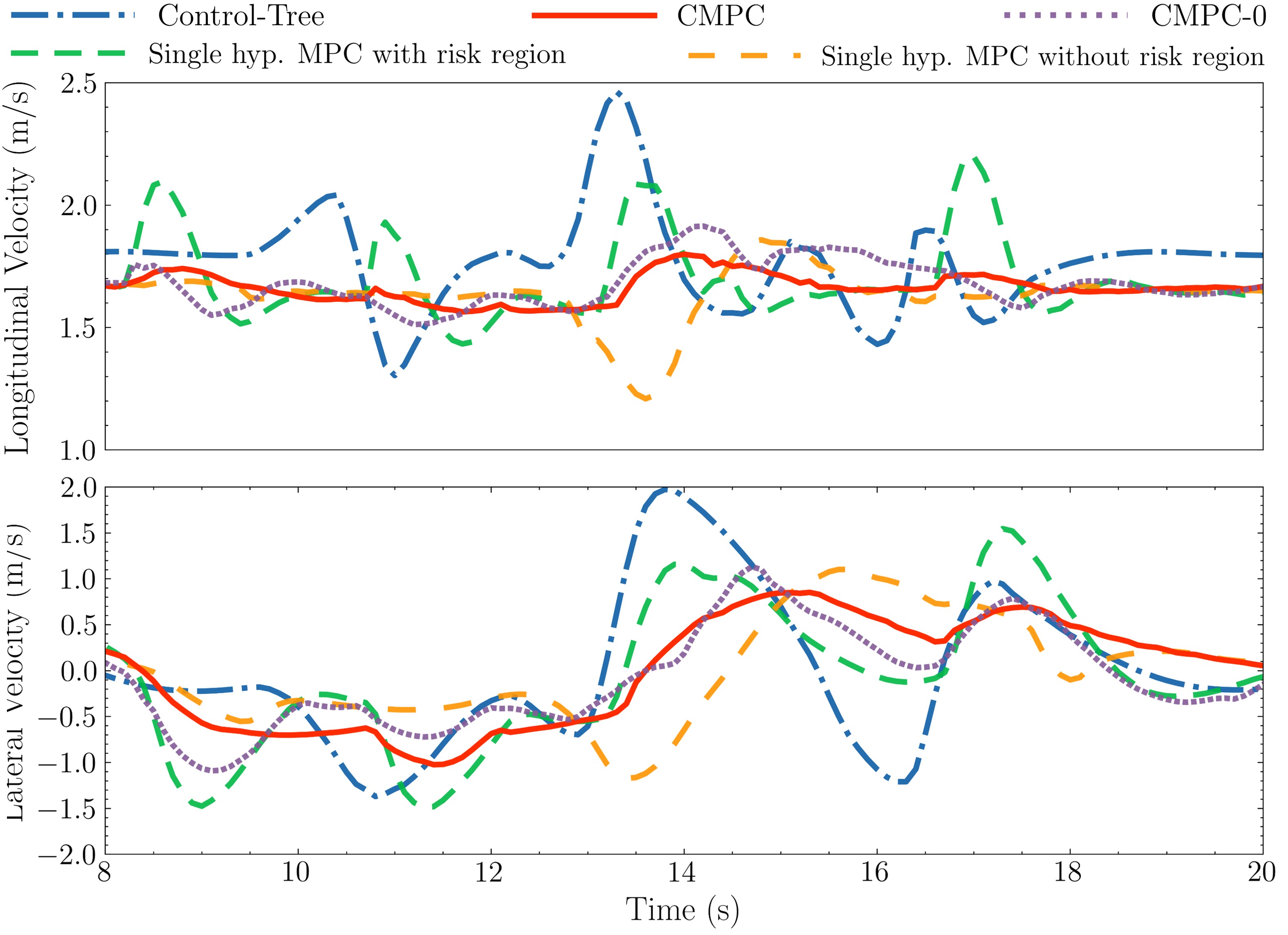} 
    \caption{The longitudinal and lateral velocity profiles of three approaches in the same scenario. When the occluded dynamic obstacles suddenly appear, both longitudinal and lateral velocities of our CMPC are more stable.} 
    \label{fig:sim_vel_comp} 
    \vspace{-0.5em}
\end{figure}

\textbf{Baseline}: We compare our approach against four baselines: the Control-Tree approach~\cite{9561664}, a distributed ADMM-based branch MPC implementation that incorporates obstacle visibility but excludes risk region considerations; a single hypothesis MPC scheme without risk region considerations; a single hypothesis MPC scheme with risk region considerations; and CMPC-0, an ablation version of CMPC without consensus segment. The first three schemes are modified from the open source code\footnote{\url{https://github.com/ControlTrees/icra2021}} of work\cite{9561664} for their best performance. The single hypothesis MPC without risk region considerations assumes the best-case scenario, while the single hypothesis MPC with risk region considerations assumes the worst-case scenario. 

To further evaluate the impact of the consensus segment, we perform an ablation study with different consensus segment lengths: $0\, \text{s}$ (no consensus), $1\, \text{s}$, $2\,\text{s}$, and $5\,\text{s}$.

\textbf{Evaluation Metrics}: Our evaluation metrics focus on three critical aspects:
\begin{itemize}
    \item \textbf{Safety Guarantee}: Whether a collision occurs. 
    \item \textbf{Motion Consistency}: Maximum lateral velocity variance and peak lateral acceleration. 
    \item \textbf{Computational Efficiency}: Average solving time. 
\end{itemize}

\subsubsection{Results}
Fig.~\ref{fig:sim_exp} shows top-down view snapshots of the simulation using our CMPC. When the robot enters the occluded, obstacle-dense region at time instant $t=8\, \text{s}$, a dynamic obstacle outlined in light-gray box is in the occlusion region blocked by Obs3, as shown in Fig.~\ref{fig:sim_exp1}. When the robot continues to go forward and approaches Obs3 at time instant $t=10\, \text{s}$, it considers the risk regions behind Obs3, as shown in Fig.~\ref{fig:sim_exp2}. The robot generates three trajectories: the dark-blue trajectory seeks the highest efficiency; the light-blue trajectory is medium aggressive, and the purple trajectory considers the worst-case to maximize safety awareness. When the dynamic obstacle is fully observable, as shown in Fig.~\ref{fig:sim_exp3}, the robot can smoothly switch to the dark-blue trajectory. 

Table~\ref{tab:comparasion_results} shows a performance comparison among different approaches. Notably, Control-Tree and the single hypothesis MPC without risk region considerations result in collisions, failing to ensure safety. A comparison of velocity profiles is shown in Fig.~\ref{fig:sim_vel_comp}. While the single hypothesis MPC with risk region considerations avoids collisions, it exhibits a 38\% higher lateral velocity variance and 51.7\% higher peak lateral acceleration compared to our CMPC. 
The ablation study comparing CMPC-0 and our approach further demonstrates that the consensus segment reduces lateral velocity variance by 15.3\% and peak lateral acceleration by 47.9\%, improving motion consistency significantly. 
Notably, the lateral velocity of these four baseline approaches increases significantly when encountering occluded dynamic obstacles, severely compromising motion consistency. 
These results indicate that our CMPC strikes a better balance between conservatism and aggressiveness, ensuring both safety and improved motion consistency. Additionally, the average solving time of our CMPC is $40.78\,\text{ms}$, indicating that it can achieve real-time trajectory generation in obstacle-dense environments. 
 
The results in Table \ref{tab:comparasion_consensus} reveal a critical trade-off when selecting the consensus segment length. A consensus segment length of $2\,\text{s}$ achieves the best performance, yielding the lowest lateral velocity variance and peak lateral acceleration compared to the shorter consensus segment lengths of $0\,\text{s}$ and $1\,\text{s}$. However, extending the consensus segment length further to $5\,\text{s}$ leads to a performance degradation in these key metrics while increasing the average solving time. This decline in performance is attributed to the ``delayed decisions''
induced by an excessively long consensus segment, which hinders the system's ability to react agilely to changes in the environment. 
Therefore, future research will explore adaptive methods for dynamically adjusting the consensus segment length based on environmental complexity and robot state to balance performance and efficiency.

 \begin{figure}[t]
 \vspace{0mm}
 \subfigcapskip = -0mm
	 \centering  
        \subfigure[The robot approaches the obstacle-dense region and considers the risk regions behind the static obstacle. 
        \label{fig:real_exp1}]{
		\includegraphics[width=0.45\columnwidth]{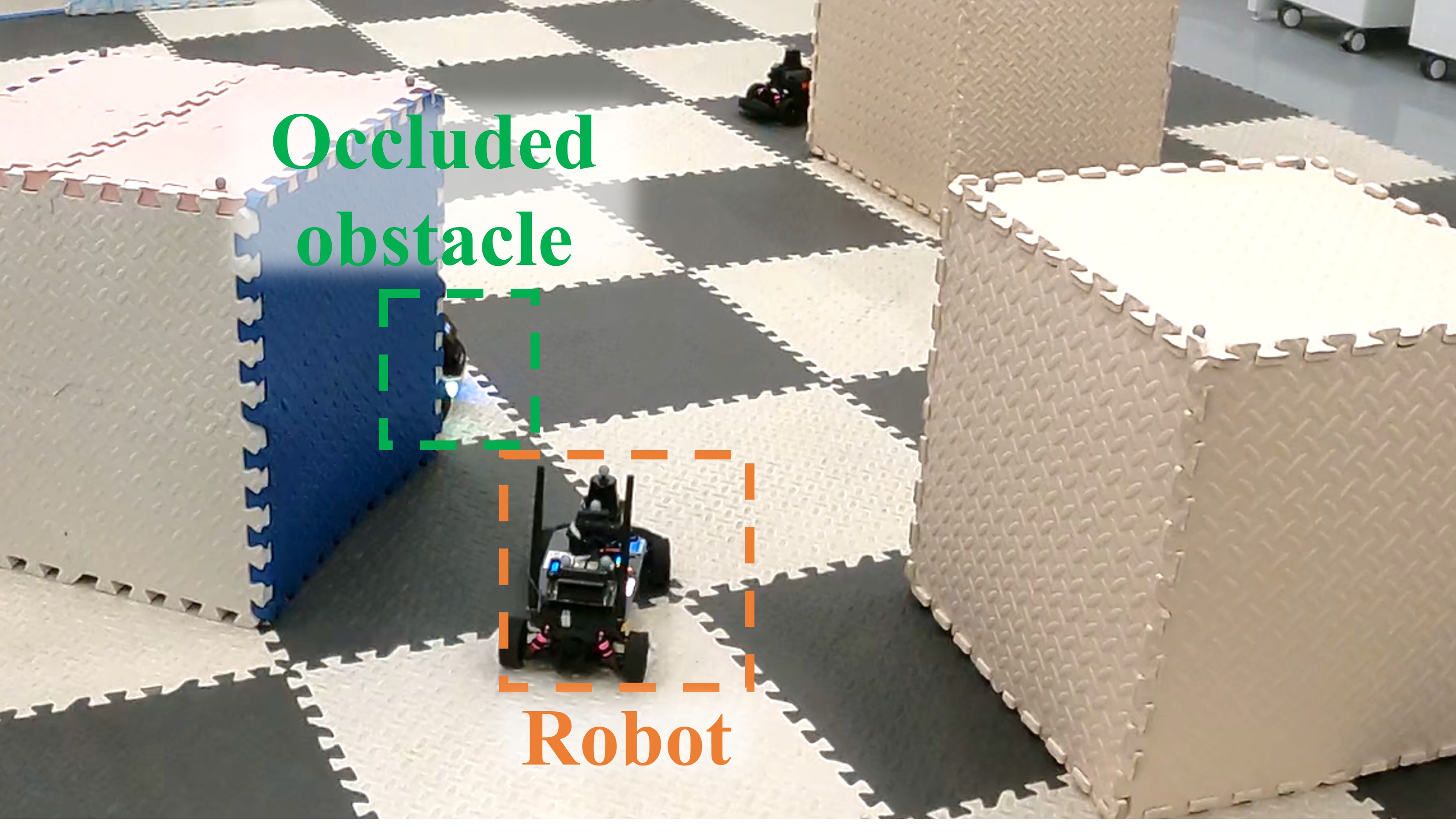}
           \hfill 
           \includegraphics[width=0.45\columnwidth]{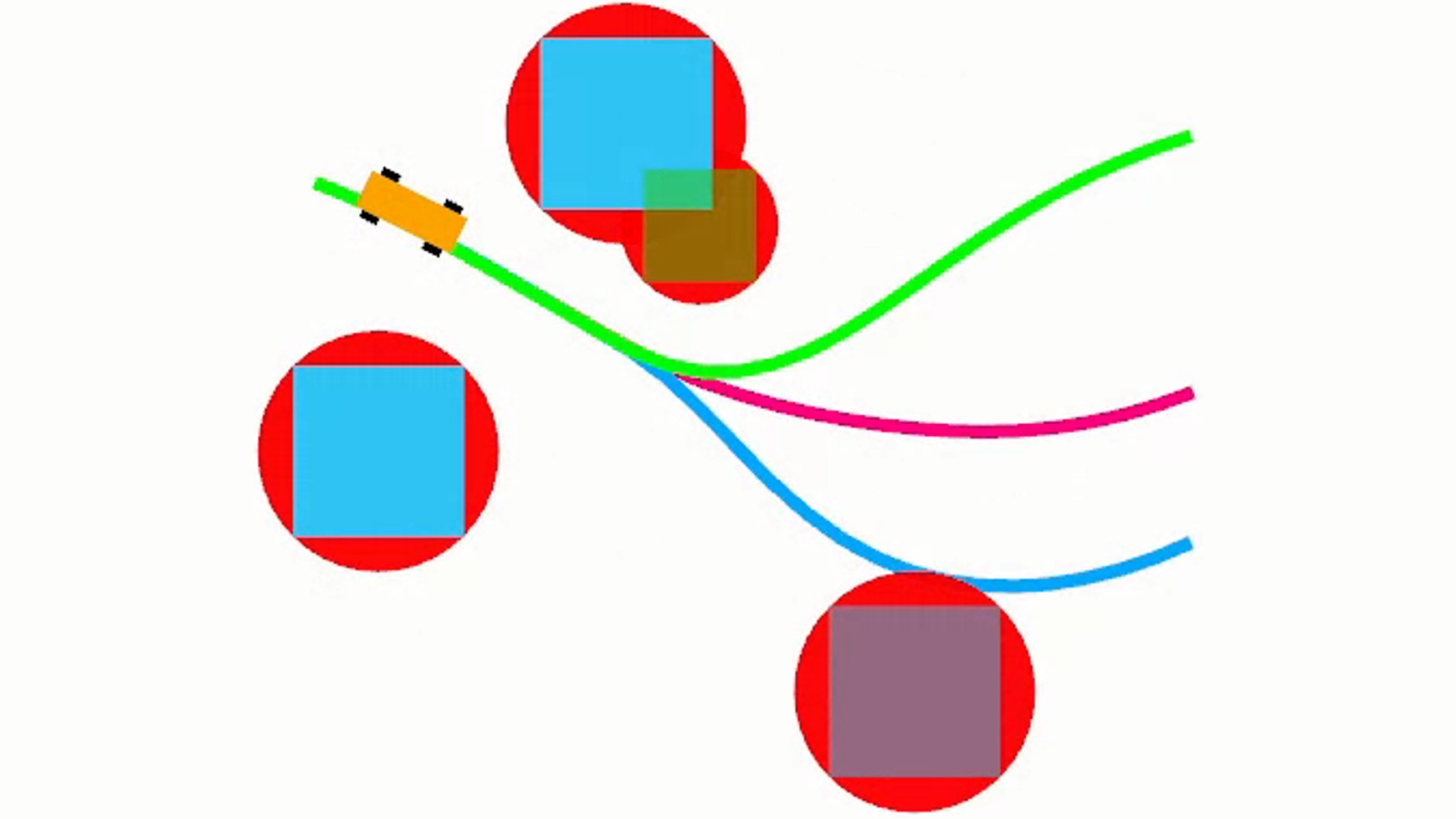}}\\
        \subfigure[The robot slightly moves away from the occluded region to avoid the potential occluded dynamic obstacle. 
        \label{fig:real_exp2}]{
		\includegraphics[width=0.45\columnwidth]{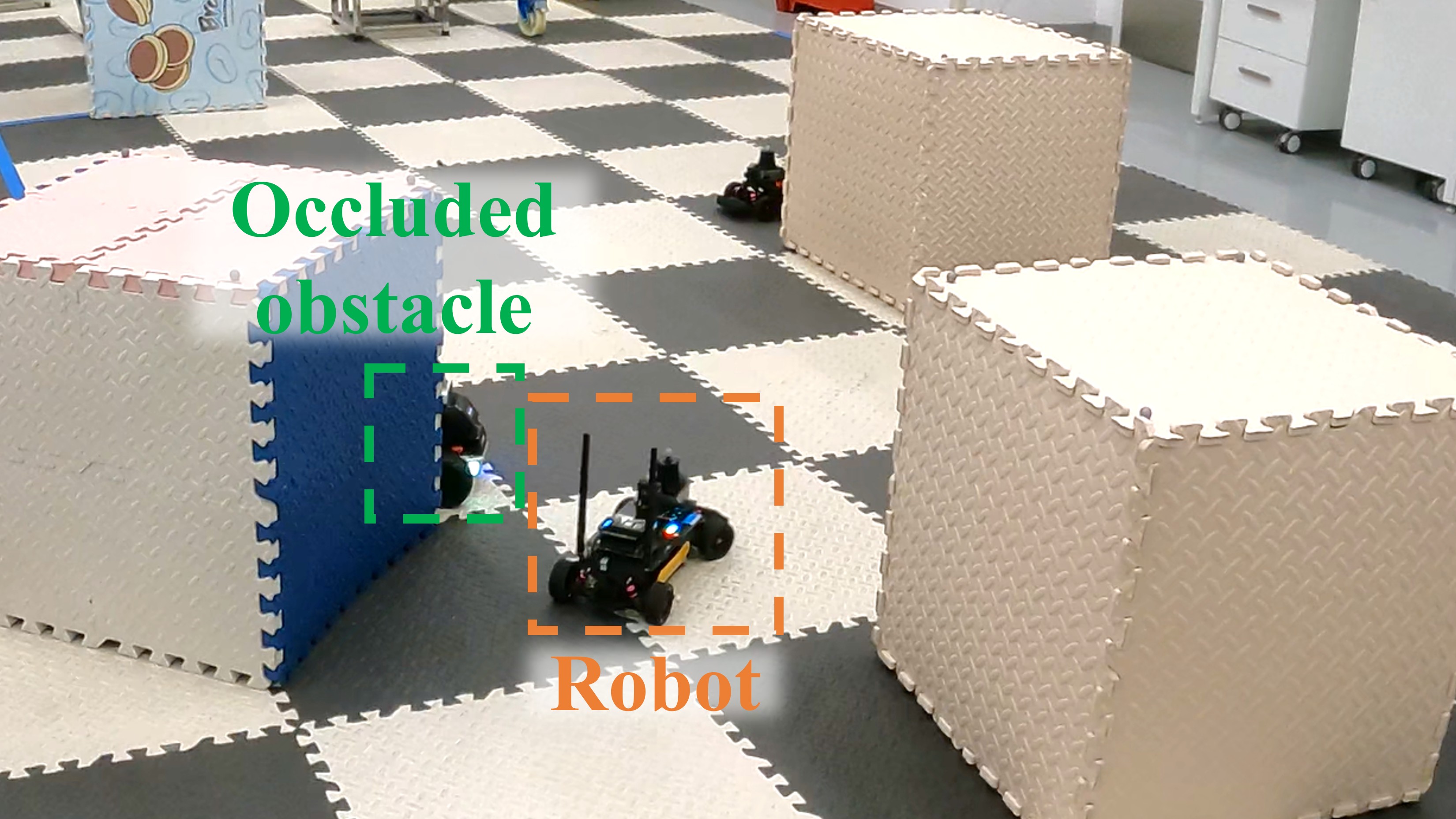}
            \hfill
        \includegraphics[width=0.45\columnwidth]{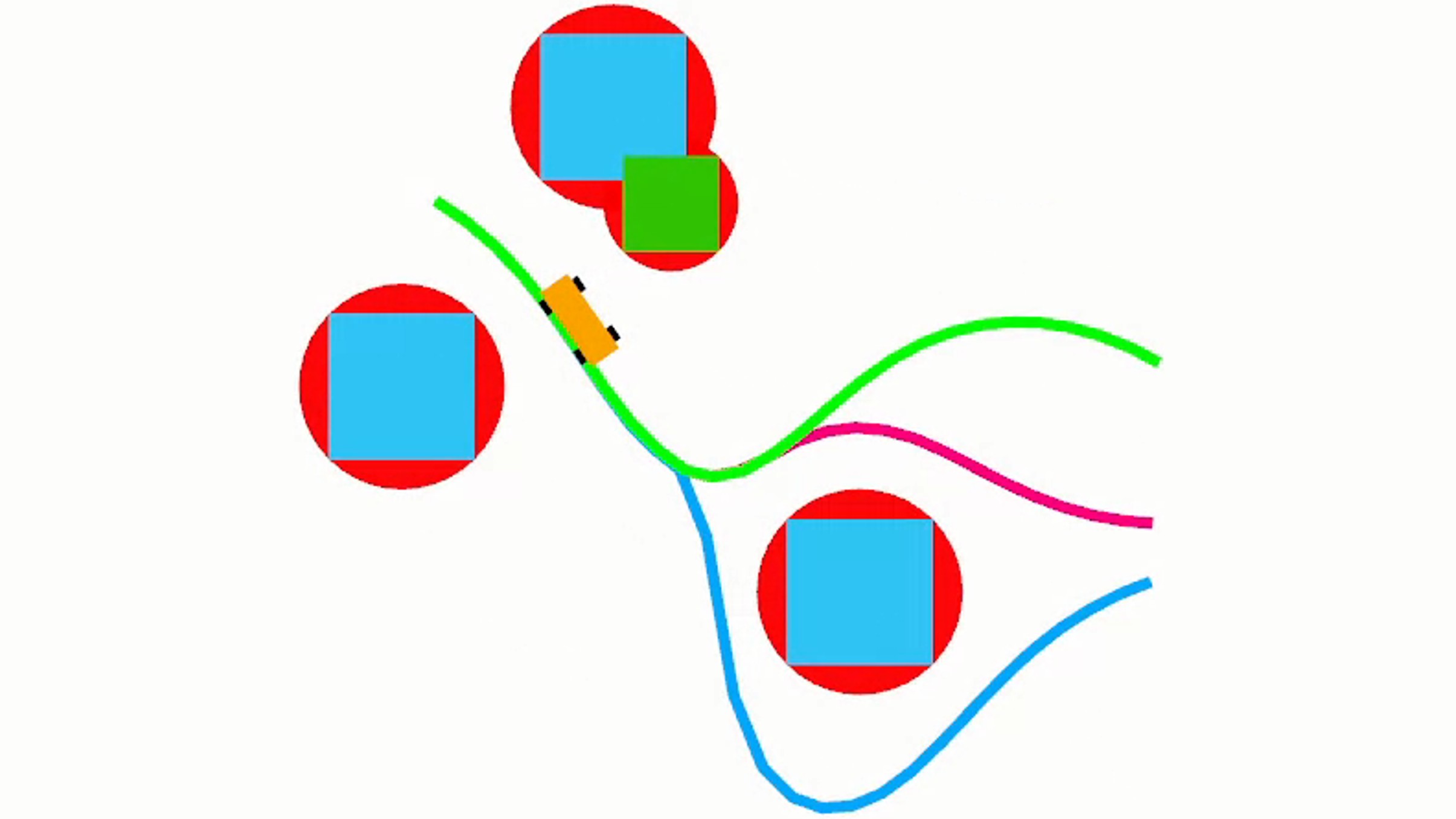}} \\
        \subfigure[The robot successfully avoids collision with the occluded dynamic obstacle. Without occluded regions, the trajectories converge to the reference path. 
        \label{fig:real_exp3}]{
		\includegraphics[width=0.45\columnwidth]{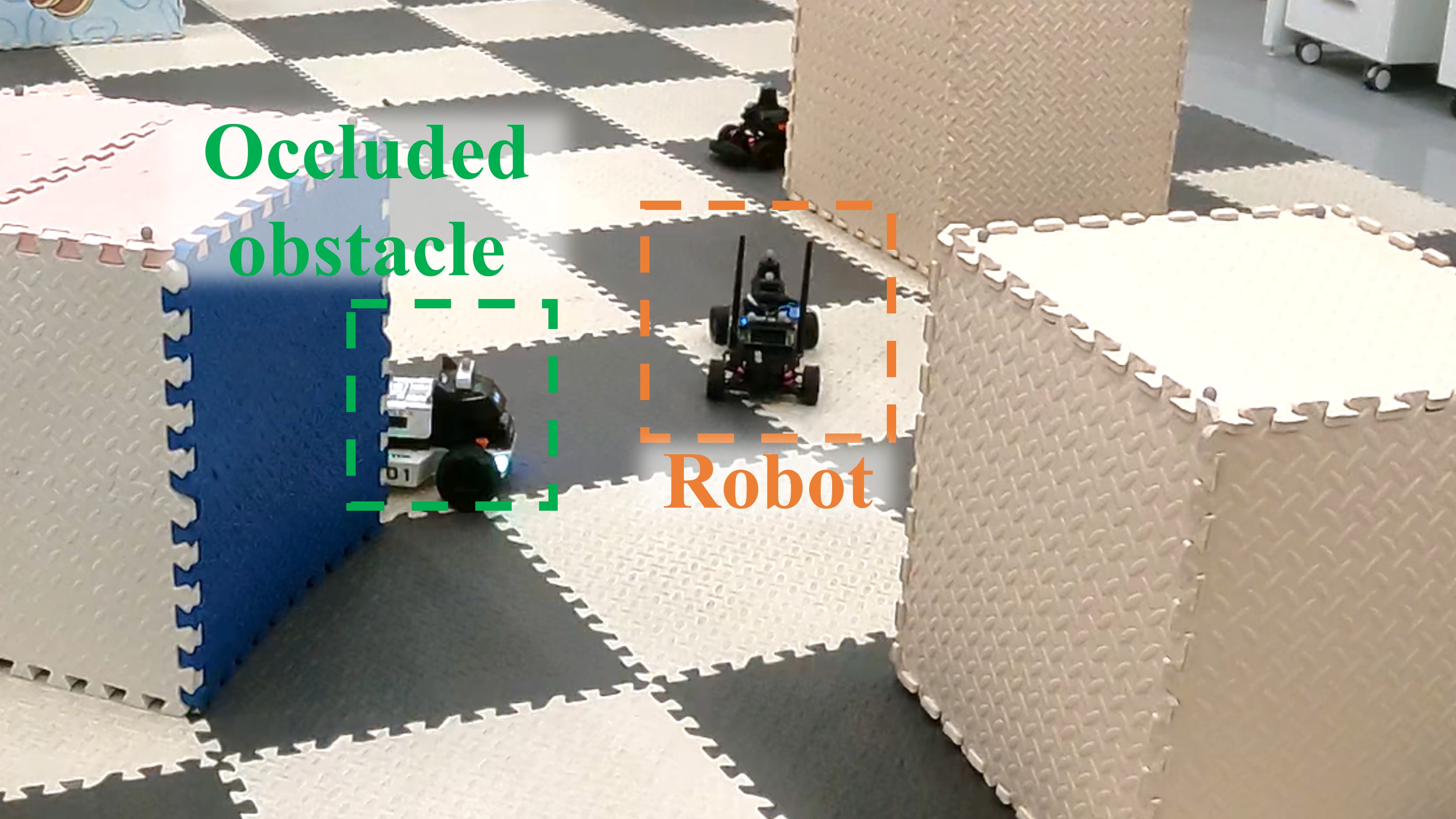}
          \hfill  \includegraphics[width=0.45\columnwidth]{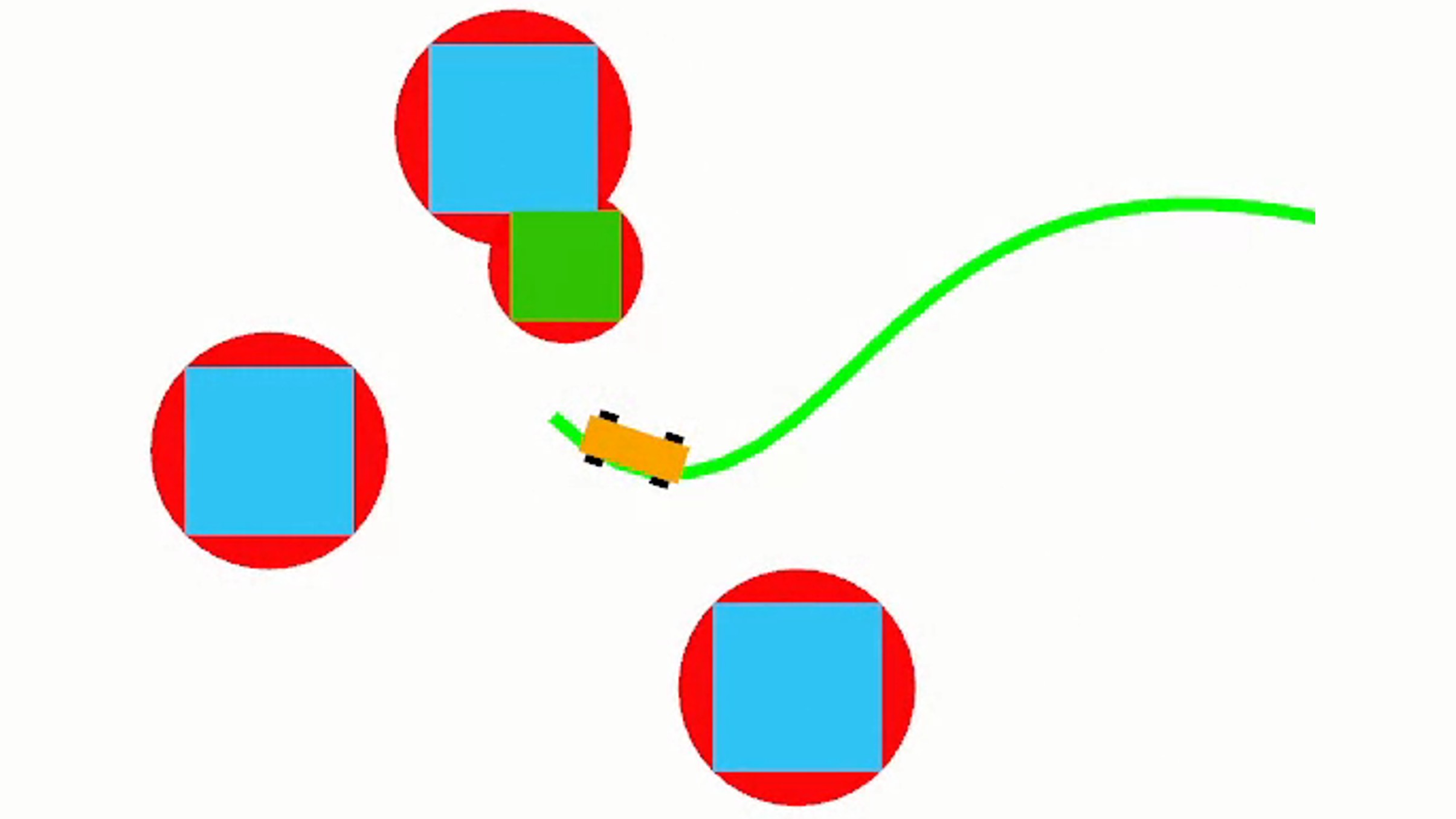}} \\
        \vspace{-1mm}
	 \caption{Real-world experiment snapshots on the TianRacer robot. The robot successfully navigates through obstacles and avoids collision with the occluded dynamic obstacle. }	
	 \label{fig:real_exp}
	 \vspace{-0.5em}
\end{figure}	

\subsection{Real-World Experiment}
\subsubsection{Experiment Setup}
The proposed CMPC strategy is deployed on a TianRacer robot, a mobile robot platform equipped with a four-wheel Ackermann-steering configuration, based on ROS1. The robot's dimensions are $380\,\text{mm} \times 210\,\text{mm}$. Static blocks sized $600 \,  \text{mm} \times 600 \,\text{mm}$ serve as static obstacles, while dynamic obstacles are represented by other mobile robots. The reference velocity of the robot is set to $0.5\,\text{m/s}$, and dynamic obstacles initiate movement at $0.2\, \text{m/s}$ when the robot is $0.5\, \text{m}$ away from them. 

\subsubsection{Results}
Fig.~\ref{fig:real_exp} shows snapshots of the TianRacer robot navigating through obstacles. It successfully avoids collisions with the suddenly appearing occluded obstacle. 

Initially, the robot follows a reference path, maintaining a safe distance from obstacles, as shown in Fig.~\ref{fig:real_exp1}. It considers the risk regions behind the obstacle and adjusts its trajectories to avoid them in the planning horizon. As the robot approaches the occluded region, it adjusts its trajectory slightly away from the occluded region to avoid potential collisions with the suddenly appearing occluded obstacle, as shown in Fig.~\ref{fig:real_exp2}. The robot continuously updates the occluded regions and plans its trajectories in response to the dynamic environment. Once the occluded region is visible, it resumes following the reference path provided by the guidance planner, as shown in Fig.~\ref{fig:real_exp3}. When there are no more occluded regions that influence the robot's movement, all branches converge to the reference path. 
 
\section{Conclusions}\label{conclusions}
In this study, we introduce a novel occlusion-aware CMPC for the safe navigation of mobile robots in occluded, obstacle-dense environments. The CMPC incorporates a module for modeling occluded regions and risk regions to proactively address potential safety threats from occluded obstacles. Coupled with a tree-structure CMPC and a common consensus segment, our strategy ensures safe traversal through occluded, obstacle-dense environments while maintaining motion consistency and task efficiency. Additionally, the integration of ADMM-based optimization enhances the computational efficiency of our strategy, enabling real-time trajectory generation. Extensive simulations demonstrate the effectiveness of our approach, achieving safe navigation with 38\% reduction in lateral velocity variance and 51.7\% lower peak lateral acceleration versus baselines. Real-world experiments on an Ackermann-steering mobile robot platform further demonstrate the effectiveness of our CMPC strategy, enabling safe navigation in physically occluded, obstacle-dense environments.
 As part of our future research, we will investigate adaptive adjustment of the consensus segment length to further balance performance and efficiency.

\bibliographystyle{IEEEtran}
\bibliography{ref}

\end{document}